\documentclass[runningheads]{llncs}

\usepackage{eccv}

\usepackage{eccvabbrv}

\usepackage{graphicx}
\usepackage{booktabs}

\usepackage{multirow}
\usepackage{bm}

\usepackage{marvosym}

\usepackage[accsupp]{axessibility}  

\usepackage{hyperref}

\usepackage{orcidlink}

\begin{document}

\title{
Mask as Supervision: Leveraging Unified Mask Information for Unsupervised 3D Pose Estimation} 

\titlerunning{Mask as Supervision}

\author{Yuchen Yang\thanks{Work performed during his internship at Shanghai Artificial Intelligence Laboratory.}\inst{1,2}\orcidlink{0009-0002-2907-6458} \and
Yu Qiao\inst{2}\orcidlink{0000-0002-1889-2567} \and
Xiao Sun\textsuperscript{\Letter}\inst{2}\orcidlink{0000-0001-7459-804X}}

\authorrunning{Y.~Yang et al.}

\institute{
Fudan University, Shanghai, China \and
Shanghai Artificial Intelligence Laboratory, Shanghai, China \\
\email{yangyc22@m.fudan.edu.cn \ \ \{qiaoyu, sunxiao\}@pjlab.org.cn}
}

\maketitle

\begin{abstract}
Automatic estimation of 3D human pose from monocular RGB images is a challenging and unsolved problem in computer vision. 
In a supervised manner, approaches heavily rely on laborious annotations and present hampered generalization ability due to the limited diversity of 3D pose datasets. 
To address these challenges, we propose a unified framework that leverages mask as supervision for unsupervised 3D pose estimation.
With general unsupervised segmentation algorithms, the proposed model employs skeleton and physique representations that exploit accurate pose information from coarse to fine.
Compared with previous unsupervised approaches, we organize the human skeleton in a fully unsupervised way which enables the processing of annotation-free data and provides ready-to-use estimation results.
Comprehensive experiments demonstrate our state-of-the-art pose estimation performance on Human3.6M and MPI-INF-3DHP datasets. Further experiments on in-the-wild datasets also illustrate the capability to access more data to boost our model. Code will be available at \url{https://github.com/Charrrrrlie/Mask-as-Supervision}.

\end{abstract}
\section{Introduction}
\label{sec:intro}

Accurate 3D pose estimation plays a pivotal role in a myriad of domains, including human-computer interaction (HCI)~\cite{liu2022arhpe, xu2019toward}, robotics~\cite{pan2023tax, hentout2019human, liu2021collision}, sports performance analysis~\cite{singh2023fast, wang2019ai}, 
and augmented/virtual reality~\cite{han2023vr,malik2020virtual}. 
However, acquiring annotated 3D data presents considerable challenges.
While there are some multi-view databases~\cite{ionescu2013human3, mono-3dhp2017, peng2021neural} in controlled environments, 
obtaining labeled 3D data remains a costly, time-consuming, and labor-intensive process. Moreover, the quantity and diversity of 3D data, both in terms of appearance and motion, are significantly inadequate. These limitations impede the generalization and robustness of 3D human pose estimation techniques when applied to novel environments.


\begin{figure}[t]
\centering
\includegraphics[width=0.9\linewidth]{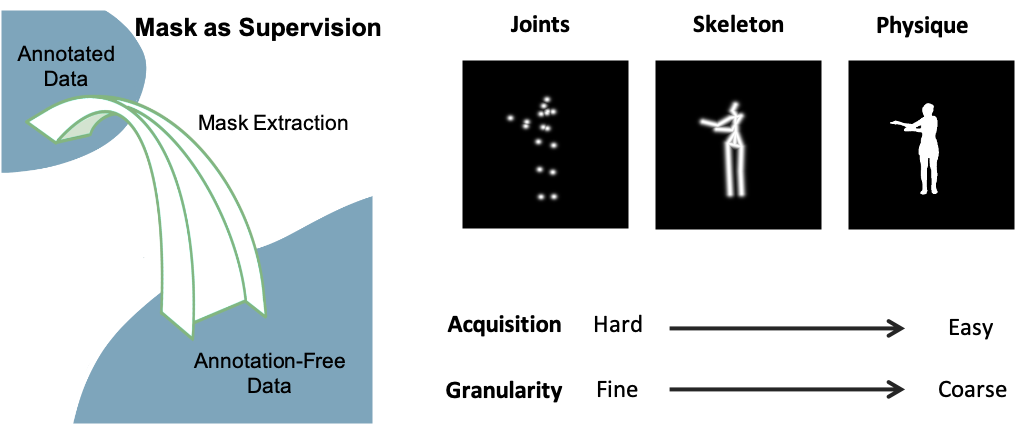}
\caption{The schematic of mask as supervision. For unsupervised pose estimation, we aim to bridge the gap of annotations and utilize attainable data. 
Meanwhile, the foreground mask is easy to acquire and implies fine-grained information that motivates us.}
\label{fig: schematic}
\end{figure}

To address the problem in 3D pose estimation, several methods attempt to predict 3D keypoints in an unsupervised manner. With no annotation required, it aims to effectively leverage vast data and impair the bias from particular human labeling to enhance the model's ability in generalization and robustness.
Generally, implicit~\cite{zhang2018unsupervised, lorenz2019unsupervised} and explicit~\cite{siarohin2019first, he2022autolink, honari2022unsupervised, sun2023bkind} human modeling from predicted keypoints are conducted. 
With background modeling or foreground masks, supervision can be provided by human reconstruction in the image. 
However, disregarding the interrelationship in human skeletons, the keypoints lack interpretability, requiring supervised post-processing (SPP) for conversion to human joints. 
Explicit modeling with human priors and geometric constraints offer feasibility in direct prediction in both 2D and 3D~\cite{jakab2018unsupervised, sosa2023self, kundu2020self, kundu2020kinematic}.
Nonetheless, existing methods require unpaired annotations or manually designed templates, where human labor is still involved.

In this paper, we introduce a novel, unified approach for unsupervised 3D pose estimation that utilizes mask information, named Mask as Supervision. As depicted in \cref{fig: schematic}, our method is motivated by the recognition that the human body's overall contour information possesses valuable insights. Particularly when extensive prior knowledge about the body shape is available, it provides rich clues for inferring 
the precise locations of 3D keypoints. 
Importantly, these informative contour cues can be obtained in an unsupervised manner by analyzing extensive video data and utilizing general segmentation algorithms.

\begin{figure*}[ht]
    \includegraphics[width=\linewidth]{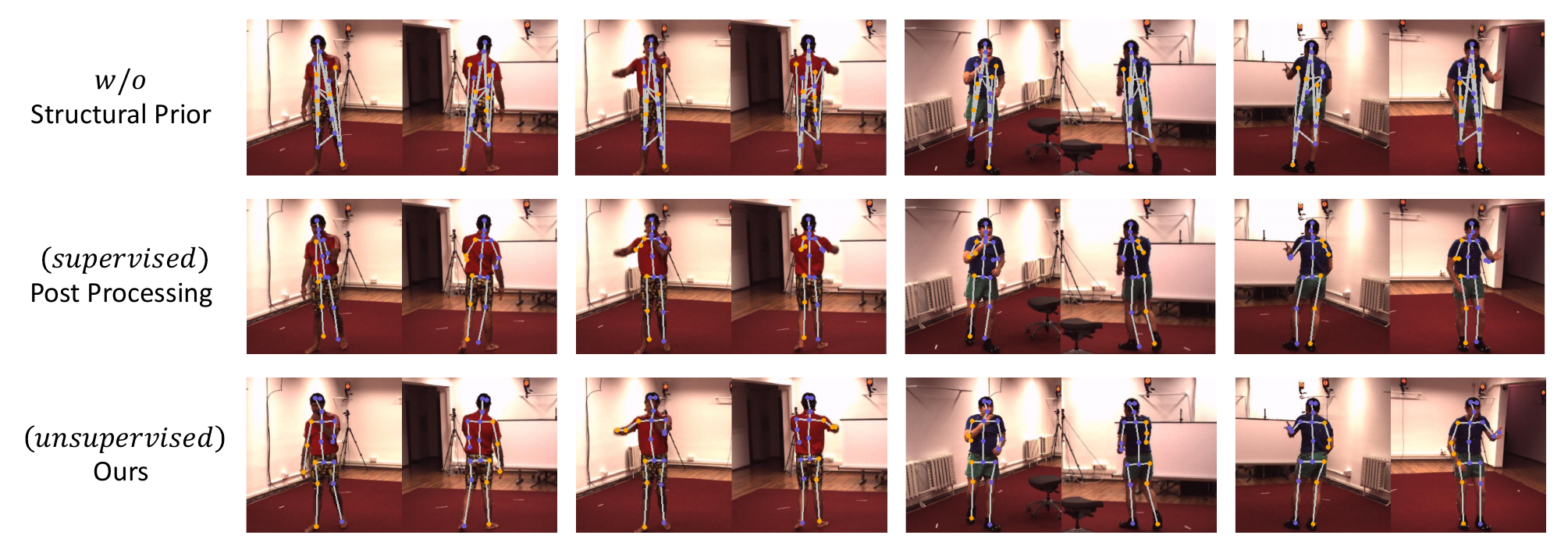}
    \caption{Different acquisition methods for structural priors. Known as an important prior, structural information leads to plausible skeletons in pose estimation. Without considering it, most previous methods necessitate supervised post-processing (SPP) during inference. To tackle this issue, we propose a method that enables effective and ready-to-use pose estimation in an unsupervised fashion.}
    \label{fig:ablation-skeleton}
\end{figure*}

To amplify the effectiveness and maximize the use of mask information, our proposed approach incorporates a range of techniques. These include a coarse-to-fine mask supervision method, defined using Skeleton and Physique Masks. 
As demonstrated in \cref{fig:ablation-skeleton}, it discovers body structure and shape priors to organize human skeletons and improve pose estimation accuracy. 
Furthermore, geodesic weighting and a cascaded optimization scheme are employed to estimate challenging distant keypoints with smooth solution space. To address the issue of left-right ambiguity, we adopt a robust strategy that effectively leverages a limited amount of unlabeled multi-view data.

Comprehensive experiments on Human3.6M~\cite{ionescu2013human3} and MPI-INF-3DHP~\cite{mono-3dhp2017} datasets demonstrate that our method achieves superior performance among state-of-the-art unsupervised pose estimation methods. 
Additionally, our model possesses the capacity to leverage in-the-wild video data without any kind of annotations. With increased training data, our model performs gradually strengthened generalization ability on unseen outdoor datasets.

Our contributions are summarised as follows:

\begin{itemize}
    \item We present a novel proxy task that employs the mask as a unique supervisory signal for unsupervised 3D pose estimation. This task empowers us to efficiently harness an extensive collection of `in the wild' data, leading to significantly enhanced performance.

    \item We introduce a set of well-established techniques, including the Skeleton and Physique Masks, Geodesic Weighting, and Cascaded Optimization. These techniques enable us to effectively extract valuable supervision from mask information by leveraging prior knowledge about the structure and shape of the human body.
    
    \item Without requiring supervised post-processing (SPP), our approach excels at predicting joints that hold physical interpretability. In the experiments, we demonstrate state-of-the-art performances in unsupervised pose estimation across widely-used datasets. 

\end{itemize}

\section{Related Work}
\label{sec:related_work}

\subsubsection{Unsupervised 2D Landmark Detection.} 
There have been plentiful methods~\cite{jakab2018unsupervised, zhang2018unsupervised, jakab2020self, siarohin2019first, lorenz2019unsupervised,he2022autolink} focusing on discovering 2D keypoints in an unsupervised fashion. 
Generic supervision is human shape modeling from keypoints to reconstruct the image foreground.
\cite{jakab2018unsupervised, zhang2018unsupervised} use Gaussian kernel to convert keypoints into heatmaps. 
\cite{lorenz2019unsupervised, jakab2020self, he2022autolink} leverage articulation prior as part representations. 
However, those predicted keypoints show no interrelationship in the human skeleton, and supervised post-processing techniques are necessitated to map them to labeled human joints.
\cite{schmidtke2021unsupervised} designed a template to explicitly explain body shape. In that way, the post-processing can be removed while the template lacks generalization ability.


\subsubsection{Unsupervised 2D to 3D Lifting.} 
Lifting methods take 2D keypoints as input and obtain 3D joints. These studies assume that the 2D aspect of the pose has been previously given which does not strictly conform to the unsupervised 3D pose estimation settings.
\cite{li2020geometry, wandt2021canonpose,kocabas2019self} constrain the network with geometric consistency by multi-view input or 3D rigid transformation.
\cite{yu2021towards, chen2019unsupervised, gong2021poseaug} utilize discrimintor to provide extra supervision from unpaired 2D keypoints.


\subsubsection{Unsupervised 3D Pose Estimation.}
Learning 3D pose without annotation is a challenge that a few existing methods attempt to address.
\cite{suwajanakorn2018discovery, sun2023bkind, honari2022unsupervised} extend 2D landmark discovery methods by leveraging multi-view constraints. Nevertheless, they still necessitate the supervised post-processing procedure.
To obtain the interpretability in keypoints, it is vital to learn effective human priors to bridle the network. \cite{kundu2020self} utilizes human-designed template and unpaired 2D keypoints to represent priors. \cite{sosa2023self} takes skeleton map as 2D representation and constraint in 3D geometry while it also demands unpaired 2D annotations.

The closest setting to our method is \cite{kundu2020kinematic}, it adopts paired images with the same background and extracts texture and shape features from raw images and keypoints. It reaches unsupervised 3D pose estimation by applying a kinematic restriction in 3D and human prior from SMPL~\cite{loper2015smpl}. However, the unpaired 3D pose prior is still necessitated and remains a gap to free the supervision from pose annotation.

As demonstrated later, our proposed method requires no pose annotations or templates and effectively organizes skeletal information for interpretable joint prediction.
\section{Method}
\label{sec:method}




\subsection{3D Pose Estimation from Single Images}
\label{sec:base_method}
We commence our discussion with a general formulation for the task of 3D human pose estimation from a single image.
Given an input image $\mathbf{I}$, our goal is to determine a set of joint locations $\mathbf{X} \in \mathbb{R}^{J \times 3}$ in the 3D coordinate system, 
\begin{equation}
    \phi(\mathbf{I}) = \mathbf{X}
\end{equation}
where $J$ represents the number of joints as defined by the particular dataset in use.
Typically, the $\phi$ network is implemented via Hourglass~\cite{newell2016stacked} as the backbone network, complemented by an underlying 3D heatmap representation. This heatmap subsequently predicts the 3D coordinates of the joints through an integral operation \cite{sun2018integral}.


\begin{figure*}[htbp]
\centering
\includegraphics[width=\linewidth]{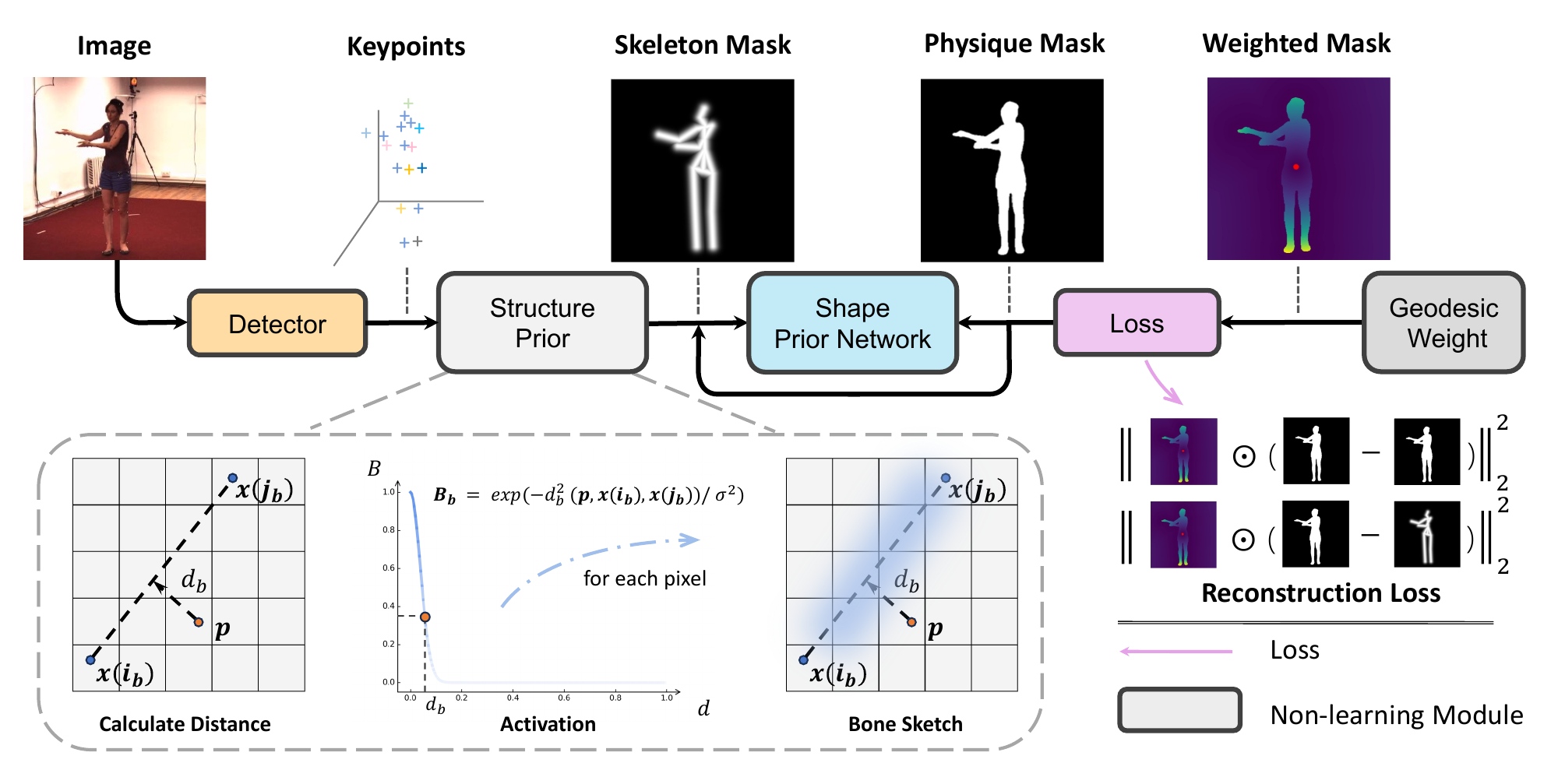}
\caption{Overview. We aim to gain supervision from mask reconstruction for the 3D detector. The Skeleton Mask and the Physique Mask representations are proposed for reconstruction in a coarse-to-refine granularity. Additionally, Geodesic Weighting is adopted to further leverage mask information. Note that only the detector will be used during inference.}
\label{fig: framework}
\end{figure*}

\subsection{Mask as Supervision}
\label{sec:mask-as-supervision}

\subsubsection
{Problem Definition.}
For supervised learning, the ground truth joint locations $\mathbf{X^{gt}}$ are provided. The backbone network is then optimized by minimizing the difference between these ground truth locations and the predicted positions, which is formally defined as the loss function $L=||\mathbf{X^{gt}}-\mathbf{X}||_2^2$.
However, in unsupervised settings, the ground truth joint locations $\mathbf{X^{gt}}$ are not available. 

Motivated by various off-the-shelf unsupervised foreground mask extraction methods~\cite{stauffer1999adaptive, boykov2006graph, lian2023bootstrapping, kirillov2023segment}, we assume that the \textbf{MASK} ground truth $\mathbf{M^{gt}}$, representative of the human form, is available and used as supervision for $\mathbf{X}$ instead.
Intuitively, one can regard $\mathbf{M^{gt}}$ as a degraded version of $\mathbf{X^{gt}}$ as depicted in \cref{fig: schematic}. 

As a result, \textbf{Mask as Supervision} can be defined as exploring a mask representation $\mathbf{Z}$ that guides the pose estimator for effective training. Without loss of generality, we denote $\mathbf{x}$ as the 2D counterpart of $\mathbf{X}$. It can be demonstrated as
\begin{equation}
    L_{Mask} = || \mathbf{M^{gt}} - \mathbf{Z}(\mathbf{x})||_2^2
\end{equation}

\subsubsection
{Baseline.}
A straightforward baseline is that all the predicted joints $\mathbf{x}$ should lie within the confines of mask $\mathbf{M^{gt}}$. 
Concurrently, the joints $\mathbf{x}$ should ideally envelop as much of $\mathbf{M^{gt}}$ as possible. This yields a baseline mask representation that uses Gaussian kernel to interpret keypoints as follows,
\begin{equation}
\label{eq:gaussian-baseline}
    \mathbf{Z}_{Base} = \mathbf{M}_{Gauss}(\mathbf{x})
\end{equation}
This simple baseline leads to a mass of ambiguities in the estimation. For example, interchanging any two keypoint locations in $\mathbf{x}$ does not result in a variation in the loss value. 

To address this issue, we propose leveraging prior knowledge concerning the human body's \textbf{\emph{Structure}} and \textbf{\emph{Shape}} to mitigate erroneous predictions.
As demonstrated in \cref{fig: framework},
this is achieved through two distinct representations: the \textbf{\emph{Skeleton Mask}} and the \textbf{\emph{Physique Mask}}, corresponding to structure and shape priors, respectively. These representations are deployed sequentially in a manner that transitions from coarse to refined granularity.

\subsubsection
{Skeleton Mask from Body Structure Prior.} Keypoints on the human body do not exist independently; rather, they establish distinct interrelationships with articulated parts that collectively define the skeletal structure. This information aids in deriving the keypoints' positions from the mask, with the consideration that the skeletal bones should also reside within the mask's boundaries.

The body structure prior is formally defined as $\mathbf{P} = \{(i_b, j_b) | b \in B\}$, where $i_b$ and $j_b$ indicate the indices of the $b$th bone's joints in $\mathbf{x}$, and there are $B$ bones in total. It constrains the geometric relationships in keypoints since only specific ones are connected as bones.

The Skeleton Mask prediction is created by drawing a differentiable line segment with a specified width for each bone. Following the methods presented in \cite{mihai2021differentiable, he2022autolink, he2023few}, we implement this using an extended Gaussian distribution across the line segment. Each bone mask, denoted as $\mathbf{B}_{b}$, is formally defined as
\begin{equation}
    \label{eq:bone-representation}
    \mathbf{B}_{b}(\mathbf{x}) = exp(-d_{b}^2(\mathbf{p}, \mathbf{x}(i_b), \mathbf{x}(j_b))/\sigma^2)
\end{equation}
where $\sigma$ is a hyper-parameter controlling bone width, and $d_{b}$ is the $L_2$ distance between pixel $\mathbf{p}$ in the map and the line segment defined by bone ($\mathbf{x}(i_b), \mathbf{x}(j_b)$). 

We then merge all bone maps via pixel-wise summation to obtain the final Skeleton Mask $\mathbf{M}_{Skel}(\mathbf{x}) = \sum_{b=1}^{B} \mathbf{B}_{b}(\mathbf{x})$. The skeleton mask loss is defined as
\begin{equation}
    L_{Skel} = || \mathbf{M^{gt}} - \mathbf{M}_{Skel}(\mathbf{x}))||_2^2
\end{equation}

As demonstrated in the experiments, the structure prior builds a connection from keypoint vectors to shape maps and effectively extracts skeletal information of the majority part within mask supervision. However, a gap still remains between the Skeleton Mask and the ground truth mask, resulting in the loss of fine details of the human body shape. The integration of a shape prior could further narrow this divergence, culminating in a more precise 3D pose estimation.


\subsubsection
{Physique Mask from Body Shape Prior.} 
Several strategies exist for the integration of effective body shape priors. 
For instance, skinned models such as SMPL~\cite{loper2015smpl} could establish the transformational relationship between the 3D keypoints $\mathbf{X}$ and the mesh vertices, and then form a more detailed Physique Mask $\mathbf{M}_{Physo}$ by projecting the mesh points onto the image. 

Even though this procedure can be rendered differentiable, their high non-linearity might complicate the system,
making it prone to settling in local optima. Instead, we propose to directly regress the physique mask from the skeleton mask via a Shape Prior Network (\textbf{SPN}), denoted as $\psi$, and find it simple yet effective.
\begin{equation}
    \mathbf{M}_{Physo}(\mathbf{x}) = \psi (\mathbf{M}_{Skel}(\mathbf{x}))
\end{equation}
where $\psi$ is implemented using a lightweight U-Net~\cite{ronneberger2015u}, and ended with a sigmoid activation~\cite{sharma2017activation} to make the pixel values in $\mathbf{M}_{Physo}$ lie in $(0,1)$. The physique mask loss is defined as
\begin{equation}
    L_{Physo} = || \mathbf{M^{gt}} - \mathbf{M}_{Physo}(\mathbf{x}))||_2^2
\end{equation}
The final objective is a weighted combination of both skeleton and physique mask losses.
\begin{equation}
\label{eq: recon_loss}
    \mathcal{L} = \lambda_s L_{Skel} + \lambda_p L_{Physo}
\end{equation}
where $\lambda_s$ and $\lambda_p$ are hyper-parameters that balance the loss weights. At the initial stage of network optimization, $\lambda_p$ is set to zero, as the $L_{Skel}$ loss function can rapidly converge and eliminate the joint ambiguities. As $L_{Skel}$ approaches convergence, the weight of $\lambda_p$ gradually increases, thereby effectively incorporating the shape prior onto the Skeleton Mask and achieving a refinement effect.

\subsubsection
{Geodesic Weighting for Hard Positives and Negatives.} 

It is well established that joints further from the root on the kinematic tree are more challenging to estimate due to the increased variation~\cite{sun2018integral}. Typically, these end joint positions appear in the mask's locations that are distant from the centroid.
Meanwhile, false positives lying on the background pixels should be unequally penalized according to the extent of deviation for smooth optimization.
These insights motivate us to enhance the relevant hard positive and negative pixels on the mask ground truth $\mathbf{M^{gt}}$. 

To this end, we employ geodesic distance~\cite{toivanen1996new} as a tool to augment the representation. The geodesic distance for an entire image, denoted as $\mathbf{G}$, can be computed using the fast marching method \cite{sethian1999fast}, based on the mask. For the foreground, we initialize the mask centroid as zero point, while for the background, all mask pixels are set to zero points. The final Geodesic Weighting map is demonstrated in \cref{fig: framework}. Consequently, \cref{eq: recon_loss} can be modified as follows:
\begin{equation}
\begin{split}
    \mathcal{L} =\lambda_s ||\mathbf{G} \odot (\mathbf{M^{gt}} - \mathbf{M}_{Skel}(\mathbf{x}))||_2^2 \\
            + \lambda_p || \mathbf{G} \odot (\mathbf{M^{gt}} - \mathbf{M}_{Physo}(\mathbf{x}))||_2^2
\end{split}
\end{equation}
Geodesic Weighting not only intensifies hard samples but also enhances the smoothness of the solution space, thus assisting in preventing convergence to local optima during the optimization process.

\subsection{Leveraging Priors in Diverse Data Modalities}
\label{sec:data-modality}
Up to this point, our framework has been discussed under the most widely applicable unsupervised single-image pose estimation setting.
Next, we will demonstrate how the ``Mask as Supervision'' approach can be effectively integrated and enhanced with various modalities of data, by capitalizing on the modality-specific prior knowledge. 


\subsubsection
{Video Data Modality.} Video is the most common visual data modality apart from images. It is comprised of a series of consecutive frames, where adjacent frames imply spatial and temporal consistency. 
Therefore, basic background modeling algorithms~\cite{boykov2006graph, reynolds2009gaussian,doretto2003dynamic} and the emerging motion-based unsupervised methods~\cite{choudhury2022guess,lian2023bootstrapping} allow us to efficiently extract the foreground mask from video data.

\subsubsection
{Multi-view Data Modality.} The presence of multi-view visual data is also increasingly noticeable in applications.
Multi-view data offers 3D geometry constraints. Specifically, given $n$ calibrated cameras,
the geometric relations of 3D joints $\mathbf{X}$ and its multi-view 2D counterparts $\mathbf{x}^\iota$ are determined via direct linear transform~\cite{hartley2003multiple} for triangulation.
\begin{equation}
\label{eq:triangulation}
    \mathbf{X} = \Delta(\{\mathbf{x}^{\iota}\}_{\iota=1}^n, \{\mathbf{\Pi}^\iota\}_{\iota=1}^n)
\end{equation}
where $\Delta$ indicates triangulation operation. $\mathbf{\Pi}^\iota =\mathbf{K}^\iota \mathbf{E}^\iota$  with camera calibration matrix $\mathbf{K},\mathbf{E}$ given.


\subsubsection
{Discussion: SPP Usage and Ambiguity in Left-Right Reversal.} Owing to the symmetry of human body, using Masks as Supervision leads to a left-right reversal problem, also mentioned in~\cite{schmidtke2021unsupervised, he2022autolink}. Networks may collapse that fail to distinguish whether a person is facing or back to the camera (i.e. all keypoints on the left of the image being treated as left joints) since both situations return the same error in the backward procedure.

In previous supervised post-processing (SPP) related works, such as our Baseline in \cref{eq:gaussian-baseline}, they can predict landmarks and some locate near to ground truth (e.g. the first line of \cref{fig:ablation-skeleton}). However, physically interpretable joints cannot be obtained since they treat joints equally and omit connectivity.
Therefore, in evaluation, SPP utilizes annotations to train a linear or two hidden-layer network $\theta: \mathbb{R}^{L \times 3} \rightarrow \mathbb{R}^{J \times 3}$ mapping high dimensional landmarks (usually $L \geq 2\times J$) to joints. 
The reversal problem is neglected with such SPP mapping. 
Worse still, it leverages training set annotations thus not fully unsupervised.

In our framework, we adopt structure and shape priors to organizing human skeletons and do not build on annotations in both training and evaluation. To tackle the left-right reversal problem, we utilize the geometry constraints in \cref{eq:triangulation} when multi-view data modality is accessible. The Skeleton Mask $\mathbf{M}_{Skel}^\iota$ is sketched by the re-projected 2D keypoints $\mathbf{x}_{prj}^{\iota} = \mathbf{\Pi}^\iota(\mathbf{X})$. It confines that different views correspond to the same 3D pose. Only by eliminating ambiguity in each view, networks can obtain correct 3D keypoints from triangulation.

\subsection{Cascaded Optimization}

\begin{figure}[ht]
    \centering
    \includegraphics[width=0.5\linewidth]{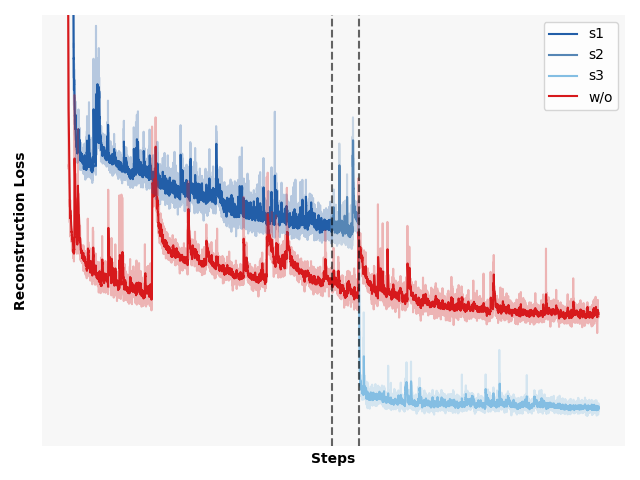}
    \caption{Reconstruction loss under different training strategies. The blue and red curves represent the cascade training and its absence, respectively. In addition, cascade stages are differentiated in colors. Losses are scaled for visualization.}
    \label{fig:cascade-loss}
\end{figure}

The human skeleton exhibits a large diversity in various actions, containing priors in high dimensionality. It leads to multiple local optima during unsupervised optimization. 

In the Mask as Supervision situation, it is simplistic to conceive how to locate the keypoints within the mask in a sub-optimal way. 
Without weakly supervised unpaired poses or laborious human-designed templates, we divide the optimization procedure into cascades with particular designs to avoid incorrect gradient decline paths. We demonstrate its effect in~\cref{fig:cascade-loss}. With the cascade strategy, the model found a shortcut in the third stage while the model erroneously fell into a local optimum without such strategy.

Concretely, we only optimize the Skeleton Mask except for arms and hands for the first stage. It helps the network find a coarse location of each joint.
As it gradually meets convergence, we then adopt the Shape Prior Network for joint optimization. Finally, arms and hands are taken into consideration, being supervised by both Skeleton and Physique Mask. 
The details can be found in the appendix.
\section{Experiments}
\label{sec:exp}

\begin{table*}[t]
  \caption{Comparison with state-of-the-art methods on Human3.6M. \textbf{SPP}: supervised post-processing. \textbf{UP}: unpaired ground truth pose or its prior, \textbf{T}: manually designed template. \textbf{SF}: supervised flip to eliminate left-right ambiguity. \textbf{\dag} indicates our results do not consider the ambiguity in left-right reversal. \textbf{\dag\dag} indicates we do not consider inner skeleton relationships and follow the common SPP settings. The best results in SPP and No SPP groups are marked in red and blue. MSE is in 2D $\%$ and MPJPEs are in $mm$.}
  \label{table:hm36-table}
  \centering
  \begin{tabular}{c|l|c|c|c|c|c|c|c|c}
    \toprule
    \multicolumn{2}{c}{\multirow{2}{*}{\textbf{Method}}} & \multicolumn{4}{|c|}{\textbf{Settings}} & \multicolumn{4}{c}{\textbf{Metrics ($\bm{\downarrow}$)}}\\
    \cmidrule{3-10}
    \multicolumn{2}{c|}{} &  {UP} & {T} & {SF} & {Joint} & {MSE} & {MPJPE} & {N-MPJPE} & {P-MPJPE} \\
    \midrule

    {\multirow{7}{*}{SPP}} & {Thewlis \textit{et al.}~\cite{thewlis2019unsupervised}} & {\color{green}{$\times$}} & {\color{green}{$\times$}} & {\color{red}{\checkmark}} & {2D} & {$7.51$} & {-} & {-} & {-} \\

    {} & {Zhang \etal~\cite{zhang2018unsupervised}} & {\color{green}{$\times$}} & {\color{green}{$\times$}} & {\color{red}{\checkmark}} & {2D} & {$4.14$} & {-} & {-} & {-} \\

    {} & {Lorenz \etal~\cite{lorenz2019unsupervised}}  & {\color{green}{$\times$}} & {\color{green}{$\times$}} & {\color{red}{\checkmark}} & {2D} & {$2.79$} & {-} & {-} & {-} \\

    {} & {Suwajanakorn \etal~\cite{suwajanakorn2018discovery}} & {\color{green}{$\times$}} & {\color{green}{$\times$}} & {\color{green}{$\times$}} & {3D} & {-} & {$158.7$} & {$156.8$} & {$112.9$} \\

    {} & {Sun \etal~\cite{sun2023bkind}} & {\color{green}{$\times$}} & {\color{green}{$\times$}} & {\color{green}{$\times$}} & {3D} & {-} & {$125.0$} & {-} & {$105.0$} \\

    {} & {Honari \etal~\cite{honari2022temporal}} & {\color{green}{$\times$}} & {\color{green}{$\times$}} & {\color{red}{\checkmark}} & {3D} & {-} & {$100.3$} & {$99.3$} & {$74.9$} \\

    {} & {Honari \etal~\cite{honari2022unsupervised}}  & {\color{green}{$\times$}} & {\color{green}{$\times$}} & {\color{green}{$\times$}} & {3D} & {\color{red}{$\mathbf{2.38}$}} & {$73.8$} & {$72.6$} & {$63.0$} \\

    \midrule
    {SPP} & {\textbf{Ours}$^{\dag\dag}$} & {\color{green}{$\times$}} & {\color{green}{$\times$}} & {\color{green}{$\times$}} & {3D} & {$2.52$} & {\color{red}{$\mathbf{65.5}$}} & {\color{red}{$\mathbf{66.1}$}} & {\color{red}{$\mathbf{61.9}$}} \\
    \midrule[0.08em]

    {\multirow{4}{*}{No SPP}} & {Schmidtke \etal~\cite{schmidtke2021unsupervised}}  & {\color{green}{$\times$}} & {\color{red}{\checkmark}} & {\color{red}{\checkmark}} & {2D} & {$3.31$} & {-} & {-} & {-} \\

    {} & {Jakab \etal~\cite{jakab2020self}} & {\color{red}{\checkmark}} & {\color{green}{$\times$}} & {\color{red}{\checkmark}} & {2D} & {\color{blue}{$\mathbf{2.73}$}} & {-} & {-} & {-} \\

    {} & {Sosa \etal~\cite{sosa2023self}} & {\color{red}{\checkmark}} & {\color{green}{$\times$}} & {\color{red}{\checkmark}} & {3D} & {-} & {-} & {-} & {$96.4$} \\

    {} & {Kundu \etal~\cite{kundu2020self}}  & {\color{red}{\checkmark}} & {\color{red}{\checkmark}} & {\color{red}{\checkmark}} & {3D} & {-} & {$99.2$} & {-} & {-} \\

    {} & {Kundu \etal~\cite{kundu2020kinematic}}  & {\color{red}{\checkmark}} & {\color{green}{$\times$}} & {\color{red}{\checkmark}} & {3D} & {-} & {-} & {-} & {$89.4$} \\
    
    \midrule
    {\multirow{2}{*}{No SPP}} & {\textbf{Ours}$^{\dag}$} & {\color{green}{$\times$}} & {\color{green}{$\times$}} & {\color{red}{\checkmark}}& {3D} & {$3.17$} & {\color{blue}{$\mathbf{85.6}$}} & {\color{blue}{$\mathbf{85.6}$}} & {\color{blue}{$\mathbf{79.3}$}} \\


    {} & {\textbf{Ours}} & {\color{green}{$\times$}} & {\color{green}{$\times$}} & {\color{green}{$\times$}} & {3D} & {$3.63$} & {$95.9$} & {$96.8$} & {$90.4$} \\

    \bottomrule
  \end{tabular}

\end{table*}

\subsection{Datasets}
We use the following datasets for training and evaluation. 
Note that, in the unsupervised setting, all ground truth poses are not accessible for our model during training. 

\textbf{Human3.6M~\cite{ionescu2013human3}.}
Following \cite{zhang2018unsupervised}, we select six activities (direction, discussion, posing, waiting, greeting, walking) and use subjects 1, 5, 6, 7, and 8 to train, subjects 9, and 11 to evaluate. 
18 joints (add thorax) are predicted and compared with ground truth.

\textbf{MPI-INF-3DHP~\cite{mono-3dhp2017}.}
Following \cite{honari2022unsupervised}, we use subjects 1 to 6 for training and subjects 7 and 8 for evaluation. We leave out the frames where the person is occluded. 
Evaluation joints are compatible with H36M.

\textbf{TikTok~\cite{jafarian2021learning}.}
It comprises single-view video sequences from social media with large appearance diversity, without any pose annotations. We select 42 sequences where most joints are visible and contain a single person.

\textbf{MPII~\cite{andriluka14cvpr}.}
Images are collected in the wild with a single view. To evaluate the generalization ability of our model, we select the validation set for experiments.


\subsection{Implementation Details}

\subsubsection{Mask Extraction.}
General segmentation methods are widely used to extract foreground masks from video data. Human3.6M dataset provides the foreground mask by the off-the-shelf graph cut algorithm~\cite{boykov2006graph}. MPI-INF-3DHP dataset adopts chroma key to extract masks in the motion capture environment with a green screen. ~\cite{remove-bg-web} is utilized in TikTok dataset. 
With the help of the multi-prompt segmentation model, we select SAM~\cite{kirillov2023segment} to fine-tune the mask with smooth and relatively accurate edges in our experiments.

\subsubsection{Training Details.} 


Experimentally, Batch Normalization~\cite{ioffe2015batch} provides inaccurate batch statistics estimation during inference, leading to poor generalization capabilities in resolving ambiguities related to left-right reversal. Hence, we adopt Group Normalization~\cite{wu2018group} to replace Batch Normalization in our detector, which avoids parameter estimation during inference. More details can be found in the appendix.

\subsection{Quantitative Analysis}

\subsubsection{Results on Human3.6M.}
In \cref{table:hm36-table}, we compare against both 2D and 3D unsupervised pose estimation methods on the Human3.6M dataset.
In previous studies, several laborious human-involved annotations or priors are introduced, which deviate from the strict definition of unsupervised learning. To clarify the experimental settings, we succinctly summarized them as follows: SPP (supervised regression in post-processing), SF (supervised flip to eliminate left-right ambiguity), UP (unpaired pose or its prior), T (manually designed template). 

Among these techniques, SPP (discussed in \cref{sec:data-modality} and \cref{sec:ablation}) relies entirely on 3D pose ground truth, which ignores the order and interrelationship of joints. Consequently, it requires separate consideration for comparison.
In contrast, UP, T, and SF avoid directly utilizing annotations to construct skeleton structures but still rely on them to guide the learning or inference process, thus we mark them up accordingly.


On the contrary, our method is able to predict interpretable 3D joints without any annotation. Fairly, we also adopt SPP to alternate the skeleton mask. We outperform most methods in 2D and 3D metrics with or without the SPP procedure, showing state-of-the-art performances.


\begin{table}[t]
  \caption{Comparison with state-of-the-art methods on MPI-INF-3DHP. MPJPE is in $cm$. Note that the first four methods use supervised post-processing and Sosa \etal \cite{sosa2023self} uses unpaired 2D pose to obtain interpretable keypoints.}

  \label{table:mpi-inf-3dhp-table}
  \centering
  \begin{tabular}{l|c|c|c}
    \toprule
    {Method} & {PCK($\uparrow$)} & {AUC($\uparrow$)} & {MPJPE($\downarrow$)} \\
    \midrule
    {Denton \etal \cite{denton2017unsupervised}} & {-} & {-} & {$22.28$} \\
    {Rhodin \etal \cite{rhodin2019neural}} & {-} & {-} & {$20.24$} \\
    {Honari \etal \cite{honari2022temporal}} & {-} & {-} & {$20.95$} \\
    {Honari \etal \cite{honari2022unsupervised}} & {-} & {-} & {$14.57$} \\
    \midrule
    {Sosa \etal \cite{sosa2023self}} & {$69.6$} & {$32.8$} & {-} \\
    \midrule
    {Ours} & {$60.2$} & {$24.7$} & {$19.36$} \\
    {Ours (SPP)} & {$\mathbf{71.3}$} & {$\mathbf{42.7}$} & {$\mathbf{13.67}$} \\
    \bottomrule

  \end{tabular}
\end{table}

\begin{figure*}[ht]
    \centering
    \includegraphics[width=\linewidth]{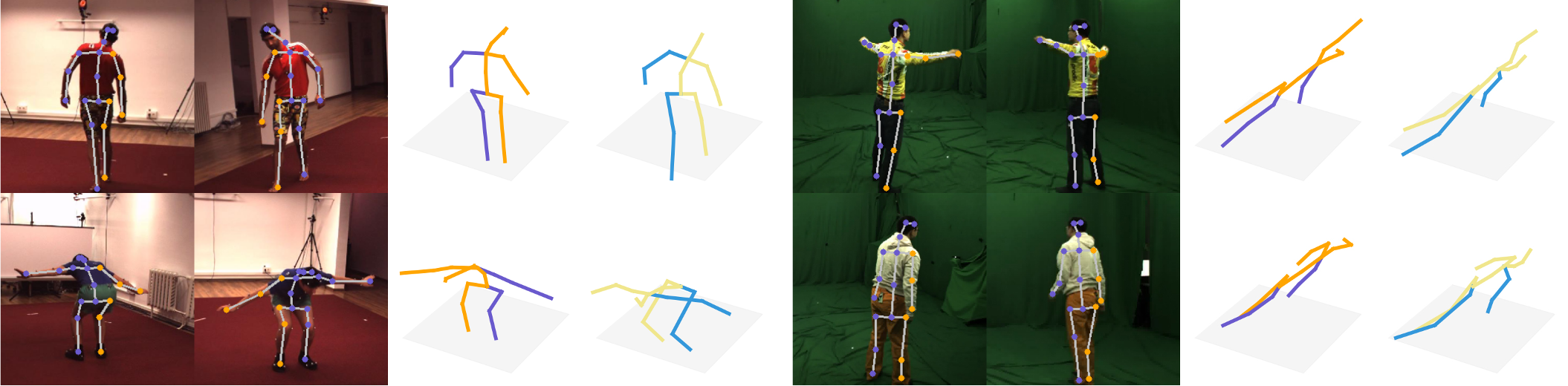}
    \caption{Qualitative results on Human3.6M and MPI-INF-3DHP datasets. We visualize joints in 2D and 3D coordinate systems, for Human3.6M (left) and MPI-INF-3DHP (right) datasets. Every $4^{th}$ column in each sub-figure shows 3D ground truth joints.}
    \label{fig:hm36-mpi-vis}
\end{figure*}

\subsubsection{Results on MPI-INF-3DHP.} To evaluate the performance on challenging scenarios, we conduct our model on MPI-INF-3DHP dataset~\cite{mono-3dhp2017}, which involves a larger diversity of actions for further exploration. \cref{table:mpi-inf-3dhp-table} presents our contestable performance with unpaired labels or SPP-necessitated methods.
Detailly, without any kind of annotations, we achieve advanced performance in MPJPE and considerable accuracy in PCK and AUC metrics. 
Note that not many unsupervised methods directly experiment on this challenging dataset. We demonstrate our results under the suggested PCK and AUC metrics in \cite{mono-3dhp2017} calling for follow-up experiments.


\subsection{Qualitative Analysis}

Qualitative results of keypoints predictions are shown in \cref{fig:hm36-mpi-vis} for Human3.6M and MPI-INF-3DHP. We address the left-right reversal problem which lies in most mask-based unsupervised methods. Through a comparison of sub-figures both within and between them, our predictions demonstrate both intra-scene and inter-scene consistency, indicating that our model has the ability to handle different people and camera views while maintaining a consistent joint definition. 


\subsection{Leveraging In-the-wild Data}

Vast amounts of data from various sources are being generated, including real-world scenarios such as video sequences captured by a single camera. As a result, it is crucial and essential for unsupervised models that can effectively learn from in-the-wild data to enhance model robustness and generalization ability. With the ability to process in-the-wild data, we firstly conduct a comprehensive analysis of the relationship between data amount and model performance in unsupervised pose estimation.

\begin{figure}[th]
    \centering
    \includegraphics[width=\linewidth]{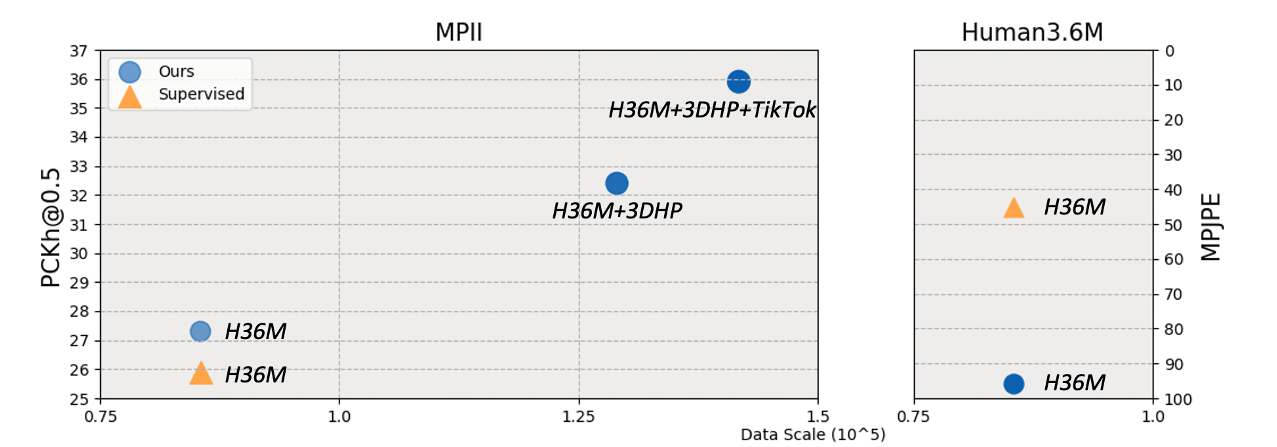}
    \caption{Scalability and generalization ability analysis. The detector is trained in mask and full supervision with training data scaling up. Sub-figures show the evaluation results on MPII and Human3.6M.}
    \label{fig:scale-up}
\end{figure}

\textbf{Scalability.} To evaluate the scalability of the proposed mask as supervision, we collect all hands-on data with mask, including Human3.6M, MPI-INF-3DHP and in-the-wild TikTok~\cite{jafarian2021learning} without any pose annotations. In \cref{fig:scale-up}, the model is trained step-by-step with scaled-up data and evaluated on the in-the-wild MPII~\cite{andriluka14cvpr}. With clear improvements, it demonstrates the capacity and scalability of our proposed method.

\begin{figure}[th]
    \centering
    \includegraphics[width=\linewidth]{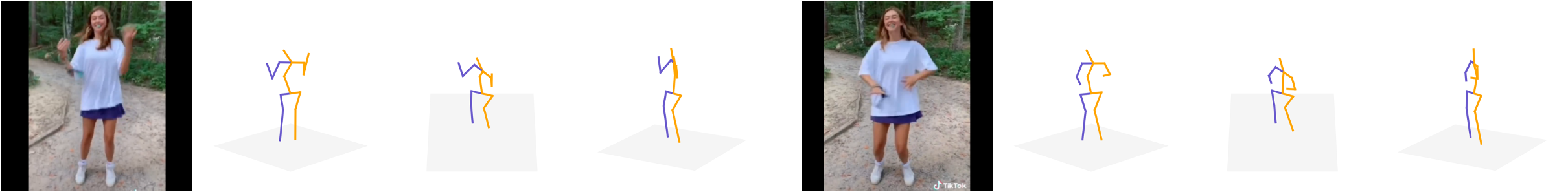}
    \caption{Qualitative results on TikTok dataset. We visualize the predictions from different views.}
    \label{fig:tiktok}
\end{figure}





\textbf{Generalization.} Additionally, the supervised can only be trained with 3D annotation when multi-view data is accessible. Keeping the detector unchanged, the fully supervised detector on in-door Human3.6M cannot generalize as effectively to in-the-wild MPII as mask supervised one (from the right sub-figure to the left).
Qualitatively, the model that leverages in-the-wild data with mask as supervision can overcome the shortages of appearance diversity in MoCap datasets. The visualization of TikTok test split is in \cref{fig:tiktok}.

\subsection{Ablation Study}
\label{sec:ablation}
\subsubsection{Skeleton Mask as Structural Representation.} To verify the effectiveness of modeling structure prior, a simple way is to select the Baseline method in \cref{eq:gaussian-baseline} that does not consider any human prior for a comparison.

As demonstrated in \cref{fig:ablation-skeleton}, the Baseline method in the first configurations fails to gain plausible skeletons due to the mass of ambiguity. Its optimization target is fitting keypoints within the mask and any interchange among keypoints will not lead to a fluctuation in the loss.
In the second row, we show the capability of the post-process technique. With SPP, skeletal structures can be organized even without any extra designs. It is powerful but not fully unsupervised thus we leverage our effective unsupervised structure prior as the replacement.

Additionally, we adopt the Shape Prior Network $\psi$ to ensure the structural prior knowledge is provided by the Skeleton Mask, rather than other proposed components. Quantitative results can be found in the appendix.

\subsubsection{Physique Mask as Shape Representation.}
We leverage the physique mask to replenish the single representation from the skeleton mask approximating the ground truth mask. Hence we design several variations to discover the functionality of the physique mask in \cref{table:ablation-generator-table}.
The Shape Prior Network $\psi$, geodesic weighting $\mathbf{G}$, and triangulation $\Delta$ are removed for baseline. Successively, those modules are rejoined into the framework for comparison. To demonstrate the results in eliminating left-right reversal ambiguity, we define an Ambiguity Ratio $r$ as a metric.

\begin{equation}
    r = \sum_{i=0}^{D} min(f_i, N-f_i) / D
\end{equation}
where $f_i$ counts the times that keypoints adding the left-right flip gain less error in a scene with $N$ camera views. $D$ indicates the number of all evaluation samples.

\begin{table}[t]
  \caption{Ablation study on shape reconstruction.}
  \label{table:ablation-generator-table}
  \centering
  \begin{tabular}{l|c|c}
    \toprule
    {Configurations} & {MPJPE ($\downarrow$)} & {Ambiguity Ratio ($\downarrow$)} \\
    \midrule
    {\textit{wo $\psi$, $\mathbf{G}$, $\Delta$}} & {$118.1$} & {$48.73\%$} \\
    {\textit{wo $\psi$, $\mathbf{G}$}}& {$127.4$} & {$23.34\%$} \\
    {\textit{wo $\mathbf{G}$}} & {$102.6$} & {$22.83\%$} \\
    \midrule
    {Full} & {$95.9$} & {$20.33\%$} \\
    \bottomrule

  \end{tabular}
\end{table}

Although without triangulation $\Delta$, the model gains better performance in MPJPE, it raises large ambiguity in left-right reversal according to our metric $r$. In summary, each component of the physique mask plays a positive role in the whole framework and can be united effectively. 

\subsection{Limitations and Discussion}
Our framework remains limited in a few aspects. Body occlusions carry ambiguity in the mask. Furthermore, inter-frame occlusions can be handled by learning from unobstructed samples while keypoints invisibility caused by frame cropping is still a challenge. Another limitation is human modeling. Currently, we consider all body parts articulated while fine-grained parts, such as the head area, do not satisfy this assumption. Upon that, fractional modeling may lead to further improvement. 
\section{Conclusion}
\label{sec:conclusion}

We present an unsupervised 3D pose estimation framework that leverages unified mask information from structure and shape representations. The proposed model effectively organizes the human skeleton with priors from the mask, making it capable of providing interpretable keypoints and accessing annotation-free data. Performance on two widely used datasets demonstrates the superior ability of our model. Moreover, we illustrate the model can be enhanced by in-the-wild video data, by virtue of the mask as supervision manner.

\newpage
\section*{Acknowledgement}
This work is partially supported by the National Key R\&D Program of China (NO.2022ZD0160100).

%
%
\bibliographystyle{splncs04}

\begin{thebibliography}{10}
\providecommand{\url}[1]{\texttt{#1}}
\providecommand{\urlprefix}{URL }
\providecommand{\doi}[1]{https://doi.org/#1}

\bibitem{remove-bg-web}
\url{https://www.remove.bg/}

\bibitem{andriluka14cvpr}
Andriluka, M., Pishchulin, L., Gehler, P., Schiele, B.: 2d human pose estimation: New benchmark and state of the art analysis. In: IEEE Conference on Computer Vision and Pattern Recognition (June 2014)

\bibitem{boykov2006graph}
Boykov, Y., Funka-Lea, G.: Graph cuts and efficient nd image segmentation. International Journal of Computer Vision  \textbf{70}(2),  109--131 (2006)

\bibitem{chen2019unsupervised}
Chen, C.H., Tyagi, A., Agrawal, A., Drover, D., Mv, R., Stojanov, S., Rehg, J.M.: Unsupervised 3d pose estimation with geometric self-supervision. In: Proceedings of the IEEE/CVF Conference on Computer Vision and Pattern Recognition. pp. 5714--5724 (2019)

\bibitem{choudhury2022guess}
Choudhury, S., Karazija, L., Laina, I., Vedaldi, A., Rupprecht, C.: Guess what moves: Unsupervised video and image segmentation by anticipating motion. In: British Machine Vision Conference (2022)

\bibitem{denton2017unsupervised}
Denton, E.L., et~al.: Unsupervised learning of disentangled representations from video. Advances in Neural Information Processing Systems  \textbf{30} (2017)

\bibitem{doretto2003dynamic}
Doretto, G., Chiuso, A., Wu, Y.N., Soatto, S.: Dynamic textures. International journal of computer vision  \textbf{51},  91--109 (2003)

\bibitem{gong2021poseaug}
Gong, K., Zhang, J., Feng, J.: Poseaug: A differentiable pose augmentation framework for 3d human pose estimation. In: Proceedings of the IEEE/CVF Conference on Computer Vision and Pattern Recognition. pp. 8575--8584 (2021)

\bibitem{han2023vr}
Han, D., Lee, R., Kim, K., Kang, H.: Vr-handnet: A visually and physically plausible hand manipulation system in virtual reality. IEEE Transactions on Visualization and Computer Graphics  (2023)

\bibitem{hartley2003multiple}
Hartley, R., Zisserman, A.: Multiple view geometry in computer vision. Cambridge University Press (2003)

\bibitem{he2023few}
He, X., Bharaj, G., Ferman, D., Rhodin, H., Garrido, P.: Few-shot geometry-aware keypoint localization. In: Proceedings of the IEEE/CVF Conference on Computer Vision and Pattern Recognition. pp. 21337--21348 (2023)

\bibitem{he2022autolink}
He, X., Wandt, B., Rhodin, H.: Autolink: Self-supervised learning of human skeletons and object outlines by linking keypoints. Advances in Neural Information Processing Systems  \textbf{35},  36123--36141 (2022)

\bibitem{hentout2019human}
Hentout, A., Aouache, M., Maoudj, A., Akli, I.: Human--robot interaction in industrial collaborative robotics: a literature review of the decade 2008--2017. Advanced Robotics  \textbf{33}(15-16),  764--799 (2019)

\bibitem{honari2022temporal}
Honari, S., Constantin, V., Rhodin, H., Salzmann, M., Fua, P.: Temporal representation learning on monocular videos for 3d human pose estimation. IEEE Transactions on Pattern Analysis and Machine Intelligence  (2022)

\bibitem{honari2022unsupervised}
Honari, S., Fua, P.: Unsupervised 3d keypoint estimation with multi-view geometry. arXiv preprint arXiv:2211.12829  (2022)

\bibitem{ioffe2015batch}
Ioffe, S., Szegedy, C.: Batch normalization: Accelerating deep network training by reducing internal covariate shift. In: International Conference on Machine Learning. pp. 448--456. PMLR (2015)

\bibitem{ionescu2013human3}
Ionescu, C., Papava, D., Olaru, V., Sminchisescu, C.: Human3. 6m: Large scale datasets and predictive methods for 3d human sensing in natural environments. IEEE Transactions on Pattern Analysis and Machine Intelligence  \textbf{36}(7),  1325--1339 (2013)

\bibitem{jafarian2021learning}
Jafarian, Y., Park, H.S.: Learning high fidelity depths of dressed humans by watching social media dance videos. In: Proceedings of the IEEE/CVF Conference on Computer Vision and Pattern Recognition. pp. 12753--12762 (2021)

\bibitem{jakab2018unsupervised}
Jakab, T., Gupta, A., Bilen, H., Vedaldi, A.: Unsupervised learning of object landmarks through conditional image generation. Advances in Neural Information Processing Systems  \textbf{31} (2018)

\bibitem{jakab2020self}
Jakab, T., Gupta, A., Bilen, H., Vedaldi, A.: Self-supervised learning of interpretable keypoints from unlabelled videos. In: Proceedings of the IEEE/CVF Conference on Computer Vision and Pattern Recognition. pp. 8787--8797 (2020)

\bibitem{kirillov2023segment}
Kirillov, A., Mintun, E., Ravi, N., Mao, H., Rolland, C., Gustafson, L., Xiao, T., Whitehead, S., Berg, A.C., Lo, W.Y., Dollar, P., Girshick, R.: Segment anything. In: Proceedings of the IEEE/CVF International Conference on Computer Vision. pp. 4015--4026 (October 2023)

\bibitem{kocabas2019self}
Kocabas, M., Karagoz, S., Akbas, E.: Self-supervised learning of 3d human pose using multi-view geometry. In: Proceedings of the IEEE/CVF Conference on Computer Vision and Pattern Recognition. pp. 1077--1086 (2019)

\bibitem{kundu2020self}
Kundu, J.N., Seth, S., Jampani, V., Rakesh, M., Babu, R.V., Chakraborty, A.: Self-supervised 3d human pose estimation via part guided novel image synthesis. In: Proceedings of the IEEE/CVF Conference on Computer Vision and Pattern Recognition. pp. 6152--6162 (2020)

\bibitem{kundu2020kinematic}
Kundu, J.N., Seth, S., Rahul, M., Rakesh, M., Radhakrishnan, V.B., Chakraborty, A.: Kinematic-structure-preserved representation for unsupervised 3d human pose estimation. In: Proceedings of the AAAI Conference on Artificial Intelligence. vol.~34, pp. 11312--11319 (2020)

\bibitem{li2020geometry}
Li, Y., Li, K., Jiang, S., Zhang, Z., Huang, C., Da~Xu, R.Y.: Geometry-driven self-supervised method for 3d human pose estimation. In: Proceedings of the AAAI Conference on Artificial Intelligence. vol.~34, pp. 11442--11449 (2020)

\bibitem{lian2023bootstrapping}
Lian, L., Wu, Z., Yu, S.X.: Bootstrapping objectness from videos by relaxed common fate and visual grouping. In: Proceedings of the IEEE/CVF Conference on Computer Vision and Pattern Recognition. pp. 14582--14591 (2023)

\bibitem{liu2022arhpe}
Liu, H., Liu, T., Zhang, Z., Sangaiah, A.K., Yang, B., Li, Y.: Arhpe: Asymmetric relation-aware representation learning for head pose estimation in industrial human--computer interaction. IEEE Transactions on Industrial Informatics  \textbf{18}(10),  7107--7117 (2022)

\bibitem{liu2021collision}
Liu, H., Wang, L.: Collision-free human-robot collaboration based on context awareness. Robotics and Computer-Integrated Manufacturing  \textbf{67},  101997 (2021)

\bibitem{loper2015smpl}
Loper, M., Mahmood, N., Romero, J., Pons-Moll, G., Black, M.J.: Smpl: A skinned multi-person linear model. ACM Transactions on Graphics  \textbf{34}(6),  1--16 (2015)

\bibitem{lorenz2019unsupervised}
Lorenz, D., Bereska, L., Milbich, T., Ommer, B.: Unsupervised part-based disentangling of object shape and appearance. In: Proceedings of the IEEE/CVF Conference on Computer Vision and Pattern Recognition. pp. 10955--10964 (2019)

\bibitem{malik2020virtual}
Malik, A.A., Masood, T., Bilberg, A.: Virtual reality in manufacturing: immersive and collaborative artificial-reality in design of human-robot workspace. International Journal of Computer Integrated Manufacturing  \textbf{33}(1),  22--37 (2020)

\bibitem{mono-3dhp2017}
Mehta, D., Rhodin, H., Casas, D., Fua, P., Sotnychenko, O., Xu, W., Theobalt, C.: Monocular 3d human pose estimation in the wild using improved cnn supervision. In: 3D Vision, 2017 Fifth International Conference on. IEEE (2017). \doi{10.1109/3dv.2017.00064}, \url{http://gvv.mpi-inf.mpg.de/3dhp_dataset}

\bibitem{mihai2021differentiable}
Mihai, D., Hare, J.: Differentiable drawing and sketching. arXiv preprint arXiv:2103.16194  (2021)

\bibitem{newell2016stacked}
Newell, A., Yang, K., Deng, J.: Stacked hourglass networks for human pose estimation. In: Computer Vision--ECCV 2016: 14th European Conference, Amsterdam, The Netherlands, October 11-14, 2016, Proceedings, Part VIII 14. pp. 483--499. Springer (2016)

\bibitem{pan2023tax}
Pan, C., Okorn, B., Zhang, H., Eisner, B., Held, D.: Tax-pose: Task-specific cross-pose estimation for robot manipulation. In: Conference on Robot Learning. pp. 1783--1792. PMLR (2023)

\bibitem{peng2021neural}
Peng, S., Zhang, Y., Xu, Y., Wang, Q., Shuai, Q., Bao, H., Zhou, X.: Neural body: Implicit neural representations with structured latent codes for novel view synthesis of dynamic humans. In: Proceedings of the IEEE/CVF Conference on Computer Vision and Pattern Recognition. pp. 9054--9063 (2021)

\bibitem{reynolds2009gaussian}
Reynolds, D.A., et~al.: Gaussian mixture models. Encyclopedia of biometrics  \textbf{741}(659-663) (2009)

\bibitem{rhodin2019neural}
Rhodin, H., Constantin, V., Katircioglu, I., Salzmann, M., Fua, P.: Neural scene decomposition for multi-person motion capture. In: Proceedings of the IEEE/CVF Conference on Computer Vision and Pattern Recognition. pp. 7703--7713 (2019)

\bibitem{ronneberger2015u}
Ronneberger, O., Fischer, P., Brox, T.: U-net: Convolutional networks for biomedical image segmentation. In: Medical Image Computing and Computer-Assisted Intervention--MICCAI 2015: 18th International Conference, Munich, Germany, October 5-9, 2015, Proceedings, Part III 18. pp. 234--241. Springer (2015)

\bibitem{schmidtke2021unsupervised}
Schmidtke, L., Vlontzos, A., Ellershaw, S., Lukens, A., Arichi, T., Kainz, B.: Unsupervised human pose estimation through transforming shape templates. In: Proceedings of the IEEE/CVF Conference on Computer Vision and Pattern Recognition. pp. 2484--2494 (2021)

\bibitem{sethian1999fast}
Sethian, J.A.: Fast marching methods. SIAM review  \textbf{41}(2),  199--235 (1999)

\bibitem{sharma2017activation}
Sharma, S., Sharma, S., Athaiya, A.: Activation functions in neural networks. Towards Data Sci  \textbf{6}(12),  310--316 (2017)

\bibitem{siarohin2019first}
Siarohin, A., Lathuili{\`e}re, S., Tulyakov, S., Ricci, E., Sebe, N.: First order motion model for image animation. Advances in Neural Information Processing Systems  \textbf{32} (2019)

\bibitem{singh2023fast}
Singh, A., Bevilacqua, A., Nguyen, T.L., Hu, F., McGuinness, K., O’Reilly, M., Whelan, D., Caulfield, B., Ifrim, G.: Fast and robust video-based exercise classification via body pose tracking and scalable multivariate time series classifiers. Data Mining and Knowledge Discovery  \textbf{37}(2),  873--912 (2023)

\bibitem{sosa2023self}
Sosa, J., Hogg, D.: Self-supervised 3d human pose estimation from a single image. In: Proceedings of the IEEE/CVF Conference on Computer Vision and Pattern Recognition. pp. 4787--4796 (2023)

\bibitem{stauffer1999adaptive}
Stauffer, C., Grimson, W.E.L.: Adaptive background mixture models for real-time tracking. In: Proceedings. 1999 IEEE Computer Society Conference on Computer Vision and Pattern Recognition (Cat. No PR00149). vol.~2, pp. 246--252. IEEE (1999)

\bibitem{sun2023bkind}
Sun, J.J., Karashchuk, L., Dravid, A., Ryou, S., Fereidooni, S., Tuthill, J.C., Katsaggelos, A., Brunton, B.W., Gkioxari, G., Kennedy, A., et~al.: Bkind-3d: Self-supervised 3d keypoint discovery from multi-view videos. In: Proceedings of the IEEE/CVF Conference on Computer Vision and Pattern Recognition. pp. 9001--9010 (2023)

\bibitem{sun2018integral}
Sun, X., Xiao, B., Wei, F., Liang, S., Wei, Y.: Integral human pose regression. In: Proceedings of the European Conference on Computer Vision. pp. 529--545 (2018)

\bibitem{suwajanakorn2018discovery}
Suwajanakorn, S., Snavely, N., Tompson, J.J., Norouzi, M.: Discovery of latent 3d keypoints via end-to-end geometric reasoning. Advances in Neural Information Processing Systems  \textbf{31} (2018)

\bibitem{thewlis2019unsupervised}
Thewlis, J., Albanie, S., Bilen, H., Vedaldi, A.: Unsupervised learning of landmarks by descriptor vector exchange. In: Proceedings of the IEEE/CVF International Conference on Computer Vision. pp. 6361--6371 (2019)

\bibitem{toivanen1996new}
Toivanen, P.J.: New geodosic distance transforms for gray-scale images. Pattern Recognition Letters  \textbf{17}(5),  437--450 (1996)

\bibitem{wandt2021canonpose}
Wandt, B., Rudolph, M., Zell, P., Rhodin, H., Rosenhahn, B.: Canonpose: Self-supervised monocular 3d human pose estimation in the wild. In: Proceedings of the IEEE/CVF Conference on Computer Vision and Pattern Recognition. pp. 13294--13304 (2021)

\bibitem{wang2019ai}
Wang, J., Qiu, K., Peng, H., Fu, J., Zhu, J.: Ai coach: Deep human pose estimation and analysis for personalized athletic training assistance. In: Proceedings of the 27th ACM International Conference on Multimedia. pp. 374--382 (2019)

\bibitem{wu2018group}
Wu, Y., He, K.: Group normalization. In: Proceedings of the European Conference on Computer Vision. pp. 3--19 (2018)

\bibitem{xu2019toward}
Xu, W.: Toward human-centered ai: a perspective from human-computer interaction. Interactions  \textbf{26}(4),  42--46 (2019)

\bibitem{yu2021towards}
Yu, Z., Ni, B., Xu, J., Wang, J., Zhao, C., Zhang, W.: Towards alleviating the modeling ambiguity of unsupervised monocular 3d human pose estimation. In: Proceedings of the IEEE/CVF International Conference on Computer Vision. pp. 8651--8660 (2021)

\bibitem{zhang2018unsupervised}
Zhang, Y., Guo, Y., Jin, Y., Luo, Y., He, Z., Lee, H.: Unsupervised discovery of object landmarks as structural representations. In: Proceedings of the IEEE Conference on Computer Vision and Pattern Recognition. pp. 2694--2703 (2018)

\end{thebibliography}

\end{document}